\documentclass{article}
\usepackage[utf8]{inputenc}
\usepackage[affil-it]{authblk}
\usepackage{hyperref}
\usepackage{bm}
\usepackage{cite}
\usepackage{nameref}
\usepackage{float}
\usepackage{courier}
\usepackage{amsmath}
\usepackage{amssymb}
\usepackage{graphicx}
\usepackage{subcaption}
\usepackage{rotating}
\usepackage{caption}
\usepackage[version=3]{mhchem} 
\usepackage{comment}
\usepackage{xcolor}
\usepackage[left=4cm, right=4cm]{geometry}
\usepackage[colorinlistoftodos]{todonotes}
\usepackage[tableposition=top]{caption}
\usepackage{microtype}

\title{Physics-Informed Generative Solver: Bridging Data-Driven Priors and Conservation Laws for Stable Spatiotemporal Field Reconstruction}
\author[1,2]{Ziyuan Zhu\textsuperscript{\#}} 
\author[1,2]{Keyu Hu\textsuperscript{\#}}
\author[1,2]{Zhifei Chen\textsuperscript{\#*}}
\author[1,2]{Yuhao Shi}
\author[1,2]{Ming Bao\textsuperscript{*}}
\author[1,2,3]{Jing Zhao}
\author[4]{Gang Wang\textsuperscript}
\author[1,5,6]{Haitan Xu}
\author[7]{Jiadong Li}
\author[8]{Qijun Zhao}
\author[9]{Xiaodong Li}
\author[1,2,10,11]{Minghui Lu\textsuperscript{*}}
\author[2,11]{Yanfeng Chen\textsuperscript}
\date{}
\affil[1]{\footnotesize School of Advanced Manufacturing Engineering, Nanjing University, Suzhou, 215163, China.}
\affil[2]{\footnotesize National Laboratory of Solid State Microstructures, Nanjing University, Nanjing, 210093, China.}
\affil[3]{\footnotesize Suzhou Acoustics Industry Technology Research Institute Co., Ltd., Suzhou, 215513, China.}
\affil[4]{\footnotesize  School of Mechanical and Electric Engineering, Soochow University, Suzhou, 215131, China.}
\affil[5]{\footnotesize  Shishan Laboratory, Nanjing University, Suzhou, 215163, China.}
\affil[6]{\footnotesize  Jiangsu Key Laboratory of Quantum Information Science and Technology, Nanjing University, Suzhou 215163, China.}
\affil[7]{\footnotesize  Key Laboratory of Nanodevices and Applications, Suzhou Institute of Nano-tech and Nano-bionics, Chinese Academy of Sciences, Suzhou, 215125, China.}
\affil[8]{\footnotesize  National Key Laboratory of Helicopter Aeromechanics, Nanjing University of Aeronautics and Astronautics, Nanjing, 210016, China.}
\affil[9]{\footnotesize  Key Laboratory of Noise and Vibration Research, Institute of Acoustics, Chinese Academy of Sciences, Beijing, 100190, China.}
\affil[10]{\footnotesize  Jiangsu Key Laboratory of Artificial Functional Materials, Nanjing, 210093, China.}
\affil[11]{\footnotesize  Collaborative Innovation Center of Advanced Microstructures, Nanjing University, Nanjing, 210093, China.}

\begin{document}

\maketitle

\maketitle

\vspace{-4.0em}
\begin{center}
    \footnotesize
    \textsuperscript{\#} These authors contributed equally to this work. \\
    \textsuperscript{*} Corresponding authors: chenzhifei@nju.edu.cn; baomingnju@nju.edu.cn; luminghui@nju.edu.cn.
\end{center}

\vspace{0em}

\begin{abstract}

Spatiotemporal field reconstruction from sparse measurements is an ill-posed fundamental inverse problem across the physical sciences, wherein purely data-driven methods often violate governing dynamics. We introduce a physics-informed generative reconstruction framework that decouples martingale-stabilized prior learning from inference-time conservation-law projection. During training, Martingale-Regularized Score Matching (MRSM) couples denoising score matching with a Score Fokker–Planck Equation constraint to enforce a reverse martingale property, yielding a dynamically stable generative prior. During inference, Physics-Informed Implicit Score Sampling (PI-ISS) projects samples toward the physical manifold by back-propagating conservation-law residuals, without embedding physical penalties into training. This separation enables flexible reconstruction from extremely sparse, incomplete, and multimodal observations while preserving physical consistency. In acoustic systems, the framework co-generates coupled pressure and particle velocity fields from sparse measurements, transforming sparse physical arrays into dense virtual arrays and suppressing spatial aliasing in undersampled localization. Beyond acoustics, the same stable prior generalizes to highly chaotic Kolmogorov flows and large-scale ERA5 meteorological fields under extreme data sparsity. By synergizing dynamical stability with physics-constrained probabilistic inference, this work establishes a rigorous and generalizable paradigm for solving high-dimensional inverse problems, bridging the gap between generative artificial intelligence and first-principles science.

\end{abstract}

\section{Introduction}

Many physical investigations are faced with a simple observational constraint: the state of a system is continuous, but measurements are sparse. For example, acoustic waves evolve over space and time, turbulent flows contain rapidly interacting vortical structures, and meteorological variables form large-scale coupled patterns across the atmosphere. In all these systems, the physical field of interest is high-dimensional and dynamically constrained, whereas the available sensors usually sample only a small fraction of the underlying state. Reconstructing a complete spatiotemporal field from such limited observations remains therefore a central inverse challenge across the physical sciences \cite{arridgeSolvingInverseProblems2019, donohoCompressedSensing2006, candesRobustUncertaintyPrinciples2006}.

Partial differential equations (PDEs) constitute the mathematical foundation for describing these physical laws. Conventional numerical methods have long driven scientific computing, but they typically require well-specified governing equations, boundary conditions, initial states, and material parameters. These requirements are difficult to satisfy when only sparse, noisy, or incomplete measurements are available. The AI-for-Science paradigm, a transformative solution to circumvent these bottlenecks, catalyzes landmark breakthroughs across quantum chemistry \cite{pfauAccurateComputationQuantum2024, gongGeneralFrameworkE3equivariant2023, liDeeplearningElectronicstructureCalculation2023}, molecular dynamics \cite{chmielaAccurateGlobalMachine2023, musaelianLearningLocalEquivariant2023, zouDeepLearningModel2023, choiPredictionTransitionState2023}, biology \cite{cuiEnzymeSpecificityPrediction2025a, abramsonAccurateStructurePrediction2024, watsonNovoDesignProtein2023, madaniLargeLanguageModels2023, moorFoundationModelsGeneralist2023}, materials science \cite{zeniGenerativeModelInorganic2025, szymanskiAutonomousLaboratoryAccelerated2023, merchantScalingDeepLearning2023, bastekInversedesignNonlinearMechanical2023, dengCHGNetPretrainedUniversal2023}, and the reconstruction of high-dimensional quantum states from sparse measurements \cite{quekAdaptiveQuantumState2021, xinLocalmeasurementbasedQuantumState2019, torlaiQuantumProcessTomography2023, zhangQuantumStateTomography2021}. Within macroscopic physical field reconstruction, physics-informed neural networks (PINNs) embed governing equations into training objectives \cite{raissiPhysicsInformedDeep2017a, raissiPhysicsInformedDeep2017}, whereas operator-learning frameworks learn mappings that approximate global solution operators \cite{luDeepONetLearningNonlinear2021, liFourierNeuralOperator2021, caoLNOLaplaceNeural2023}.

Despite these advances, real-world spatiotemporal inverse problems remain fundamentally limited by observational sparsity \cite{arridgeSolvingInverseProblems2019, ongieDeepLearningTechniques2020}. We herein operationally define sparsity as a regime where the available observational degrees of freedom are profoundly insufficient to uniquely determine the target physical manifold via the Shannon-Nyquist sampling theorem or traditional well-posed equation solvers. This regime includes severe spatial downsampling, extensive random masking, and the complete absence of coupled cross-modal variables. Under such conditions, the inverse problem is intrinsically non-unique: multiple physically plausible fields may be consistent with the same sparse measurements. Deterministic neural architectures, which map a given input to a single output, are therefore prone to stability degradation and unrecoverable error accumulation in highly sparse and noisy regimes \cite{wangEigenvectorBiasFourier2021, wangUnderstandingMitigatingGradient2020}. In doing so, they often suppress rather than represent the uncertainty inherent to the inverse problem.

As a promising alternative, generative paradigms, such as score-based diffusion models \cite{songGenerativeModelingEstimating2020, songScoreBasedGenerativeModeling2021, hoDenoisingDiffusionProbabilistic2020, kingmaVariationalDiffusionModels2023, songDenoisingDiffusionImplicit2022} and flow matching models \cite{liuFlowStraightFast2022, lipmanFlowMatchingGenerative2023, albergoStochasticInterpolantsUnifying2023}, offer a powerful pathway to capture complex data distributions and resolve the intrinsic uncertainty of inverse problems \cite{kawarSNIPSSolvingNoisy2021, kawarDenoisingDiffusionRestoration2022, chungDiffusionPosteriorSampling2022, liLearningSpatiotemporalDynamics2024}. However, directly applying these models to physical systems introduces a different failure mode. A generated field may appear statistically plausible while violating conservation laws, wave propagation constraints, or momentum balance. Purely data-driven priors rely on empirical correlations and do not guarantee conservative probability flows during generation. As sparsity becomes severe, small score-estimation errors can be amplified along the reverse sampling trajectory, driving the generated samples away from the physical manifold. The key challenge is therefore not only to generate plausible fields, but to generate fields that remain dynamically stable and physically admissible under extreme data sparsity.

Here we introduce Martingale-Regularized Score Matching (MRSM), a generative framework for reconstructing spatiotemporal physical fields from sparse observations. MRSM reformulates ill-posed physical inversion as a regularized joint optimization problem. During pre-training, it couples denoising score matching (DSM) \cite{songScoreBasedGenerativeModeling2021} with Score Fokker--Planck Equation (Score FPE) regularization \cite{laiFPDiffusionImprovingScorebased2023, laiEquivalenceConsistencyTypeModels2023a}. This regularization encourages the diffusion trajectories to satisfy a reverse martingale property, thereby stabilizing the learned generative prior before any task-specific physical constraints are imposed. To enable efficient inference, we further derive an Implicit Score Sampler (ISS) tailored for variance-exploding (VE) stochastic differential equations (SDEs). For systems with explicit physical coupling, ISS is extended to a Physics-Informed ISS (PI-ISS). Rather than imposing PDE residuals during training, PI-ISS injects conservation residuals into the reverse sampling iterations as gradient constraints. This design separates prior learning from physical enforcement: the data-driven prior is learned once, while conservation laws are enforced only at inference. As a result, the method avoids the multi-objective optimization pathologies of conventional physics-informed training and supports zero-shot generalization across irregular and unstructured sparse observation patterns.

This work primarily contributes a physically constrained generative reconstruction framework that remains effective when observations are extremely sparse, incomplete, and multimodal. In acoustic systems, we validate PI-ISS using in-house acoustic vector sensors (AVS) in both standing-wave tube and free-field environments. By enforcing mass and momentum conservation, PI-ISS achieves self-consistent co-generation of acoustic pressure and particle velocity fields from sparse measurements. This converts sparse physical arrays into dense virtual arrays and suppresses the sidelobe interference that traditionally limits undersampled array signal processing. Beyond acoustics, the same MRSM prior maintains energetic and mechanical fidelity in highly chaotic Kolmogorov flows and large-scale ERA5 meteorological fields \cite{hersbach2020era5}. Together, these results suggest a shift from hardware-dense sensing toward physics-informed generative reconstruction, where sparse measurements are computationally lifted into continuous, dynamically consistent physical fields. By linking stochastic generative priors with first-principles physical constraints, this study establishes a general paradigm for multimodal inverse problems in high-dimensional physical systems.

\section{Results}

In this section, we first introduce the conceptual mechanisms of the MRSM framework and the PI-ISS strategy for solving ill-posed spatiotemporal inverse problems. Subsequently, the efficacy and generalizability of the proposed framework are evaluated across a hierarchy of dynamical systems: from idealized acoustic simulations to real-world cross-modal inferences. We then demonstrate the framework’s capability to break the spatial Nyquist limit in free-field source localization. Finally, this paradigm culminates in generalizations to highly chaotic Kolmogorov flows and meteorological systems.

\subsection{The MRSM and PI-ISS methods}

\begin{figure}
\centering
\makebox[\textwidth][c]{\includegraphics[width=1\textwidth, trim=4 0 4 0, clip]{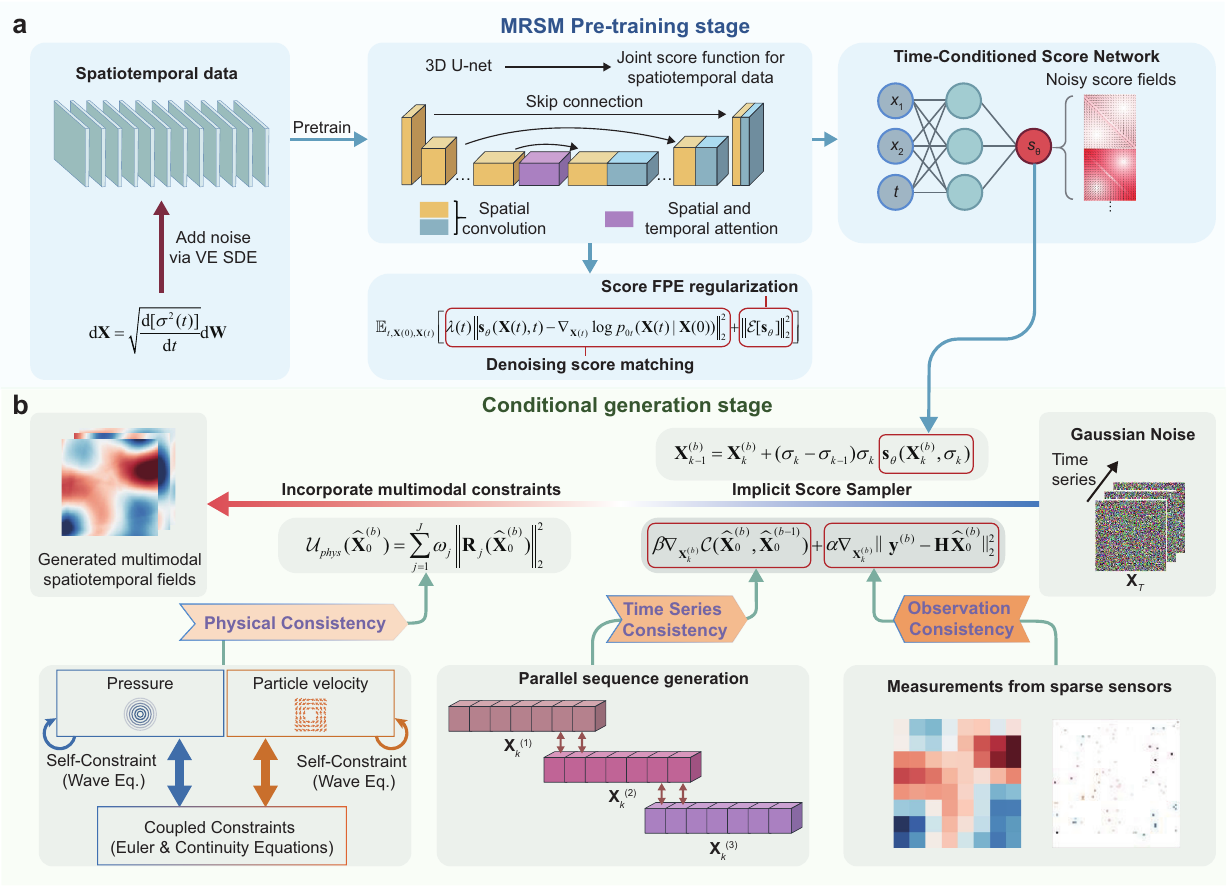}}
\caption{\textbf{Schematic of the proposed generative framework.} 
        \textbf{a}, MRSM pre-training stage. The model employs a spatiotemporal attention-enhanced neural architecture to fit the joint score function of multimodal data. The training objective synergizes DSM with Score FPE regularization to enforce dynamical stability. 
        \textbf{b}, Conditional generation stage. Iterative denoising is executed via the proposed ISS. The inference process is governed by three gradient guidance mechanisms: (1) Time Series Consistency for continuous evolution across sequential patches, (2) Observation Consistency to strictly honor sparse sensor measurements, and (3) Physical Consistency, which forces the co-generated multimodal fields to satisfy governing laws. }
\label{fig:architecture}
\end{figure}

The proposed generative framework comprises two stages: the pre-training phase enforcing martingale dynamics regularization, and the conditional generation phase driven by physics-informed guidance. The overarching conceptual architecture is illustrated in Figure \ref{fig:architecture}.

\begin{figure}
\centering
\makebox[\textwidth][c]{\includegraphics[width=1\textwidth, trim=0 0 30mm 0, clip]{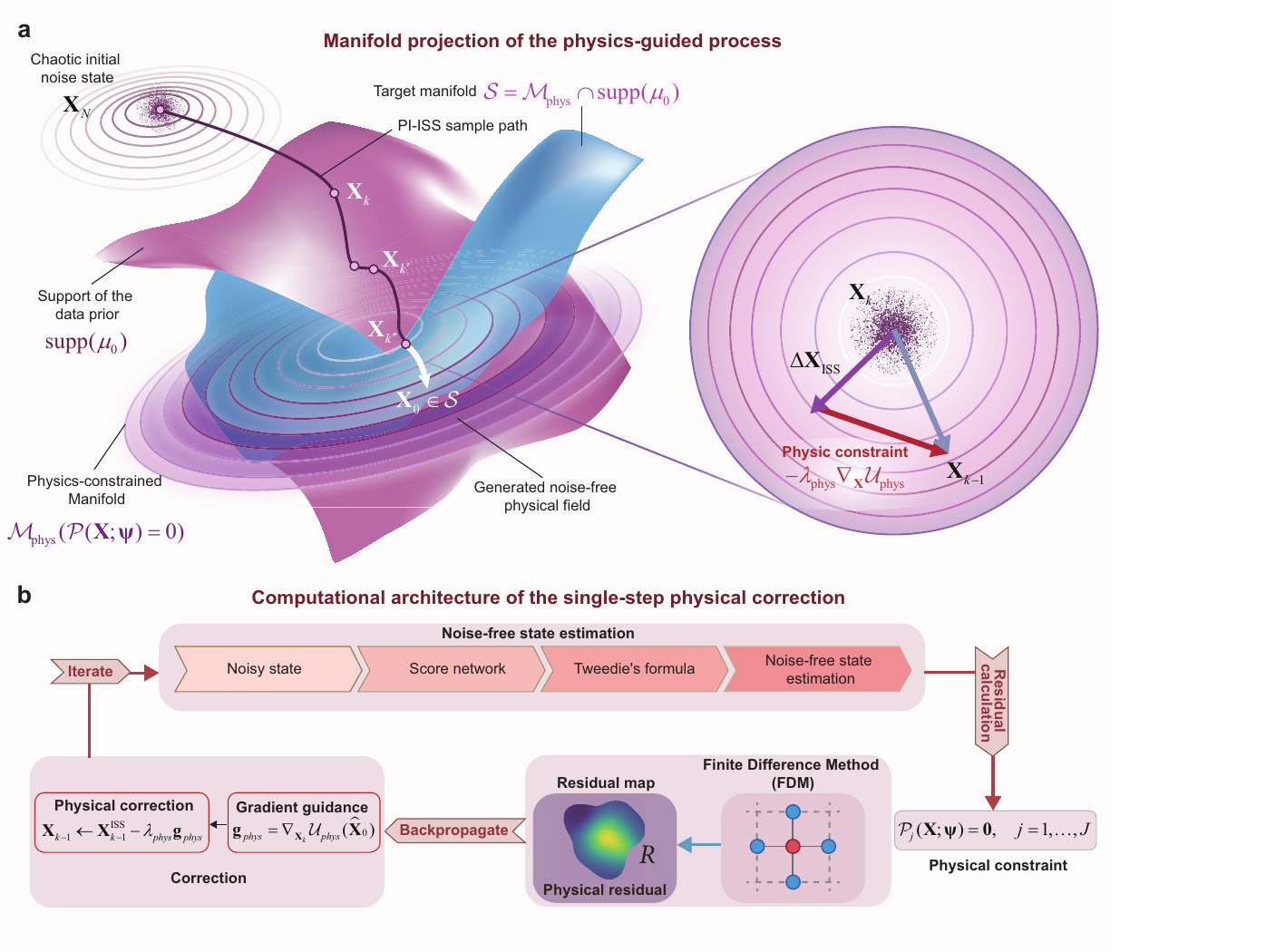}}
\caption{\textbf{Physics-informed manifold projection and gradient guidance.} 
        \textbf{a}, Left panel: manifold projection of the physics-guided process. The PI-ISS sampling trajectory originates from a chaotic noise state and is iteratively guided toward the target manifold, defined as the intersection of the data prior support and the physics-constrained manifold. Right panel: a detailed illustration of the discrete vector update, combining the unconditional ISS step with the physical gradient correction.
        \textbf{b}, Computational architecture of the single-step physical correction. Illustration of the physical correction loop (PI-ISS) during a single reverse inference step. A noise-free physical state is first estimated from the current noisy state. Physical residuals are evaluated via discrete operators, yielding the total physical potential energy. The gradient of this energy with respect to the current state, is then applied as a direct correction to the ISS update trajectory, scaled by the guidance strength.}
\label{fig:PI-ISS}
\end{figure}

In the pre-training stage, the framework learns the joint score function of the multimodal physical fields. In conventional DSM, a critical limitation is its inaccurate score estimation in low-data-density regions, leading to non-conservative generation trajectories. To circumvent this, MRSM introduces a Score FPE regularization term to the training objective. This dual-constraint mechanism goes beyond merely fitting the empirical data distribution; it constrains the evolved score field to satisfy the underlying probability flow of the forward diffusion process. This ensures that the score function itself acts as a conserved quantity, which mathematically enforces the reverse martingale property. Consequently, this formulation establishes a dynamically stable generative prior for ill-posed physical inverse problems (Figure \ref{fig:architecture}a).

In the conditional generation stage, we derive an ISS to overcome the computational bottlenecks of high-dimensional spatiotemporal sampling. This sampler, mathematically established in the Supplementary Note S1, accelerates generation while rigorously preserving strict integration accuracy. We upgrade this sampler into a PI-ISS to ensure the generated multimodal fields strictly obey physical laws. At each iterative step, PI-ISS conceptually estimates the underlying noise-free state from the current noisy trajectory, and then injects a physics-guided correction term directly into the state update rule (Figure \ref{fig:PI-ISS}b) by computing the gradient of the physical conservation residuals with respect to this estimated state. Ultimately, this cross-gradient coupling mechanism ensures asymptotic physical consistency, strictly projecting the data-driven sampling sequence onto the defined physical manifold (Figure \ref{fig:PI-ISS}a). The detailed algorithmic implementations and theoretical proofs for these mechanisms are provided in the Methods and Supplementary Note S2.

\subsection{Physics-consistent reconstruction of acoustic fields}

\begin{figure}
\centering
\makebox[\textwidth][c]{\includegraphics[width=\textwidth]{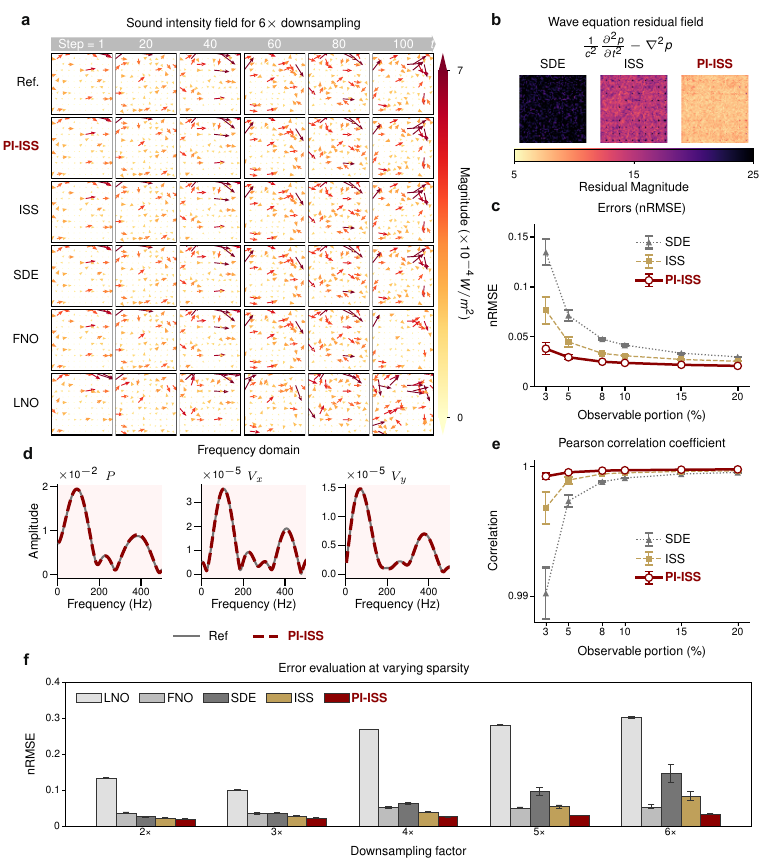}}
\caption{\textbf{Performance evaluation of free-field reconstruction.} 
        \textbf{a}, Reconstructed sound intensity fields under $6\times$ spatial downsampling. From top to bottom: Ground truth (Ref.), PI-ISS, ISS, standard SDE solver, FNO, and LNO. The baseline operator networks were trained under $6\times$ spatial downsampling.
        \textbf{b}, Physical consistency evaluated via wave equation residuals. 
        \textbf{c}, Evolution of the mean nRMSE for acoustic pressure and particle velocity components as a function of the observable data proportion.
        \textbf{d}, Frequency-domain analysis comparing the average spectrum of the PI-ISS reconstructed signals (blue solid line) against the reference (red dashed line). 
        \textbf{e}, Average Pearson correlation coefficients between the reconstructed fields and the ground truth across varying observable proportions. 
        \textbf{f}, Reconstruction performance (nRMSE) across baseline methods at varying spatial downsampling factors. For \textbf{c}, \textbf{e}, and \textbf{f}, data are presented as mean values $\pm$ standard deviation (s.d.) over 5 independent test samples.}
\label{fig:2d_free_field}
\end{figure}

To evaluate the performance of the MRSM framework in solving multimodal physical inverse problems, we first benchmarked the model on a free-field acoustic propagation task. This scenario, governed by the linearized Euler equations and the law of mass conservation, features intricate cross-modal coupling between acoustic pressure and particle velocity, and tests the physical representation capacity of generative models. To establish the generative prior, we constructed a comprehensive spatiotemporal dataset via finite-difference simulations to capture diverse wavefield dynamics excited by acoustic point sources within a non-reflecting domain (see Supplementary Note S10 for dataset details). For inference, we challenged the trained frameworks with a highly ill-posed task: reconstructing the full-field dynamics of an independent test sample from $6\times$ spatially downsampled observations corrupted by Gaussian noise to simulate sensor uncertainties.

Furthermore, to demonstrate inference efficiency, we constrained all score-based solvers (standard SDE, ISS, and PI-ISS) to a rapid 100-step generation regime, representing an order-of-magnitude reduction from the 1000-step baseline typical of standard reverse diffusion. Meanwhile, we evaluated end-to-end operator learning models, namely the Fourier Neural Operator (FNO) and Laplace Neural Operator (LNO), trained under the same  $6\times$ downsampling conditions. Under this scenario, Figure \ref{fig:2d_free_field}a illustrates the recovered sound intensity fields. While all methods successfully capture the macroscopic wavefield topology, the standard SDE solver, the ISS, and the operator networks exhibit perceptible directional deviations and subtle local distortions. In contrast, PI-ISS more faithfully recovers the intricate spatiotemporal patterns of the ground truth.

The advantage of PI-ISS becomes more evident when we examine the intrinsic adherence of the reconstructed fields to physical laws rather than mere visual similarity. While conventional statistical models can easily achieve pixel-level resemblance, they fail to guarantee mechanical consistency, frequently masking unphysical generative hallucinations. Analysis of the wave equation residuals (Figure \ref{fig:2d_free_field}b) directly exposes this vulnerability: standard SDE trajectories severely violate governing physical dynamics, a direct consequence of the large truncation errors induced by the accelerated 100-step generation regime. Although ISS is specifically formulated to overcome this rapid sampling degradation and considerably mitigates these residual levels compared to the standard SDE, it still operates without explicit physical constraints. In contrast, the physics-informed gradient guidance in PI-ISS mathematically bounds the generative trajectory within the physically permissible manifold, profoundly suppressing fundamental conservation violations and ensuring strict mechanical validity.

Quantitative evaluations confirm this advantage across varying data sparsity levels. While visual differences in the intensity fields may appear nuanced, the objective metrics clearly delineate the performance hierarchy. Notably, the standard SDE method suffers severe accuracy degradation under the 100-step constraint, whereas the proposed ISS maintains high fidelity, validating its design for efficient, accelerated sampling without requiring the standard 1000-step computational cost. Building upon this baseline, PI-ISS establishes a commanding lead. It consistently achieves the lowest normalized root-mean-square error (nRMSE) and the highest Pearson correlation for both acoustic pressure and particle velocity components (Figures \ref{fig:2d_free_field}c, e). Crucially, even in the extreme low-observation regime, PI-ISS maintains robust reconstruction fidelity.

Beyond spatial metrics, capturing the true dynamical behavior of the acoustic field requires overcoming the spectral bias of generative models, a tendency that often smooths out fine-scale dynamics or introduces unphysical high-frequency noise. Frequency-domain analysis (Figure \ref{fig:2d_free_field}d) rigorously verifies that PI-ISS overcomes this limitation. It accurately preserves the time-varying spectral properties of the acoustic signals and captures exact resonance peaks and spectral roll-off characteristics, without introducing spurious high-frequency energy blowups. This corroborates that the model has deeply internalized the acoustic wave propagation mechanisms rather than merely fitting spatial contours. Finally, PI-ISS sustains superior accuracy when benchmarked against the FNO and LNO baselines across all spatial downsampling factors (Figure \ref{fig:2d_free_field}f). These results establish that integrating data-driven generative priors with explicit physical constraints can effectively overcome the severe ill-posedness of physical inverse problems.

\subsection{Sim-to-real cross-modal inference in acoustic environments}

\begin{figure}
\centering
\makebox[\textwidth][c]{\includegraphics[width=0.95\textwidth]{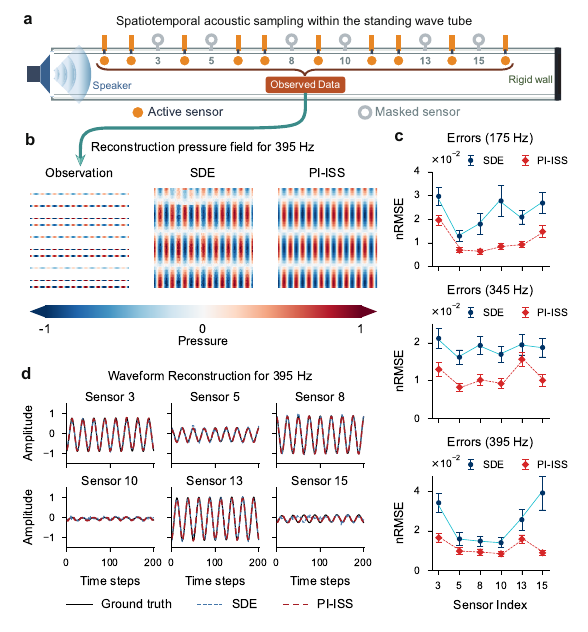}}
\caption{
        \textbf{Experimental results of the scalar acoustic pressure field.} 
        \textbf{a}, Spatiotemporal acoustic sampling within the standing wave tube. A loudspeaker at the left end generates the incident waves, and a rigid wall at the right end provides total reflection to establish the standing wave field. The setup utilizes 10 observation points as sparse inputs while withholding the remaining 6 for independent validation. 
        \textbf{b}, Comparison of spatiotemporal acoustic pressure fields. The left panel displays the sparse 10-channel observation input, while the two right panels detail the 64-node high-resolution fields reconstructed by the standard SDE and PI-ISS solvers. 
        \textbf{c}, Comparison of the nRMSE at various sensor locations across different operating frequencies. 
        \textbf{d}, Waveform comparison at validation points, contrasting the signals reconstructed by different methods at the 6 masked sensor locations against the measured ground truth.
    }
\label{fig:scalar_reconstruction}
\end{figure}

To validate the efficacy of the proposed framework in a real-world physical environment, we constructed a standing wave tube platform to perform scalar and cross-modal acoustic field reconstructions. A fundamental challenge in physical deep learning is the acquisition of dense, high-fidelity ground-truth data for training. To preclude this, the MRSM framework was pre-trained exclusively on a numerical simulation dataset governed by the 1D linearized Euler equations (see Supplementary Note S11 for simulation parameters). The subsequent inference tasks were directly executed on real-world measurements acquired via an in-house developed synchronous scalar and vector acoustic data acquisition system (detailed in Extended Data Figure \ref{fig:extended_data_1} and Supplementary Note S5), testing the model's capacity to overcome the sim-to-real domain gap.

We first conducted a sparse reconstruction experiment for the scalar acoustic pressure field, using a masked validation strategy (Figure \ref{fig:scalar_reconstruction}a). The physical measurement domain was discretized into a virtual reconstruction grid comprising 64 uniformly distributed nodes with a spatial resolution of 20 mm. During inference, a non-uniform masking scheme was applied: data from 10 of the 16 sensors were retained as sparse conditional inputs, while measurements from the remaining 6 sensors (indices 3, 5, 8, 10, 13, and 15) were strictly withheld to serve as independent validation targets.

As depicted in Figure \ref{fig:scalar_reconstruction}b for an operating frequency of 395 Hz, the absence of explicit physical constraints causes the standard SDE solver to generate unphysical spatial oscillations within the unobserved intervals. Conversely, the PI-ISS method recovers a spatially continuous and smooth standing wave profile across the entire 64-node domain. Quantitative analysis (Figure \ref{fig:scalar_reconstruction}c) confirms PI-ISS consistently achieves a lower nRMSE than the standard SDE at all sensor locations, an advantage across multiple frequencies (175 Hz, 345 Hz, and 395 Hz). Detailed waveform comparisons at the masked locations (Figure \ref{fig:scalar_reconstruction}d) further reveal that the PI-ISS reconstructions align precisely with the empirical ground truth while the standard SDE exhibits significant phase and magnitude deviations. This successful sparse reconstruction confirms the framework's robust adaptability to the inherent distributional shifts between idealized simulated training data and noisy physical measurements.

\begin{figure}
\centering
\makebox[\textwidth][c]{\includegraphics[width=\textwidth]{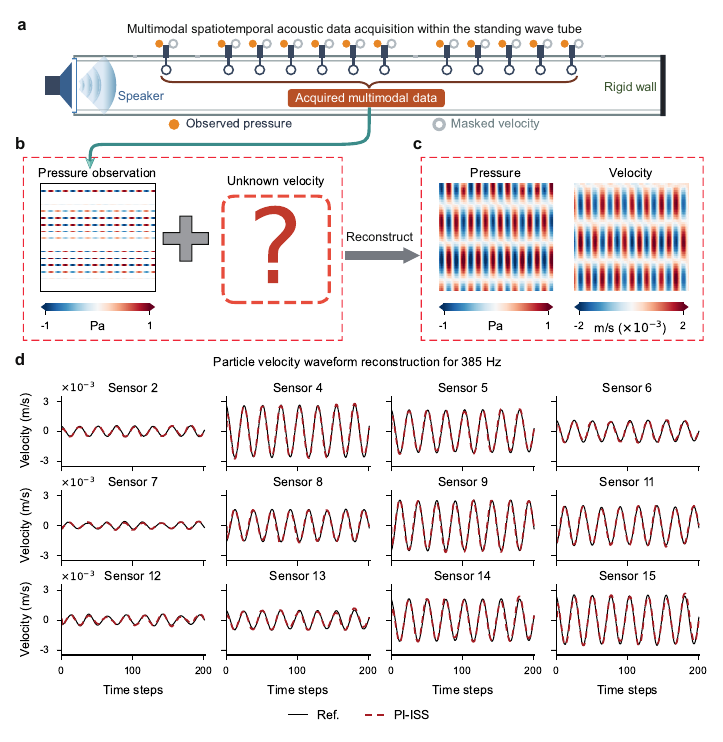}}
\caption{
        \textbf{Experimental results of cross-modal inference.} 
        \textbf{a}, Multimodal spatiotemporal acoustic data acquisition within the standing wave tube. Multimodal data are acquired using a PU joint sensor array; however, only the acoustic pressure (orange solid dots) is observed as the conditional input, while the acquired particle velocity channel (gray hollow circles) is strictly masked for cross-modal validation. 
        \textbf{b}, Schematic of the inference scheme, highlighting that the model is conditioned exclusively on pressure observations while the velocity channel is fully masked.
        \textbf{c}, Reconstructed full-field physical dynamics, showcasing the high-resolution (64 nodes) spatiotemporal evolution of acoustic pressure and particle velocity synchronously generated by PI-ISS. 
        \textbf{d}, Comparison of the particle velocity waveform reconstructions. The plots contrast the unobserved velocity waveforms inferred by PI-ISS (red dashed lines) against the measured ground truth (black solid lines) at the 12 active sensor locations under a 385 Hz excitation.
    }
\label{fig:vector_inference}
\end{figure}

Following the scalar validation, we challenged the framework with a highly ill-posed cross-modal inference task to evaluate its capability in resolving multiphysics coupled inverse problems. To acquire the requisite multimodal data, we designed and fabricated a PU (pressure-velocity) joint sensor chip in-house, specifically engineered for this validation, to synchronously measure both physical quantities (Extended Data Figure \ref{fig:extended_data_1}). Using experimental data acquired at 385 Hz, the model was conditioned exclusively on acoustic pressure observations from 12 active sensor locations, while the corresponding particle velocity channel was completely masked (Figures \ref{fig:vector_inference}a, b). The objective was to blindly infer the full-field (64 nodes) spatiotemporal distribution of both acoustic pressure and particle velocity (Figure \ref{fig:vector_inference}c).

Figure \ref{fig:vector_inference}d presents a time-domain evaluation of the particle velocity waveforms reconstructed by PI-ISS against the measured ground truth at the 12 sensor locations. PI-ISS bridges this modality gap by coupling the data-driven joint score prior with the governing laws of mass and momentum conservation, even in the complete absence of velocity observations. Specifically, the DSM captures the intrinsic statistical correlations between pressure and velocity, while the physics-informed gradient guidance mechanism mathematically translates the pressure-gradient information into physical corrections for the unobserved particle velocity. The resulting reconstructions closely match the empirical data in both magnitude and phase, corroborating that this framework can effectively solve complex cross-modal physical inverse problems by anchoring data-driven generative priors to fundamental physical laws.

\subsection{Generative virtual arrays suppress spatial aliasing}

\begin{figure}
\centering
\makebox[\textwidth][c]{\includegraphics[width=\textwidth, trim=2mm 0 0 0, clip]{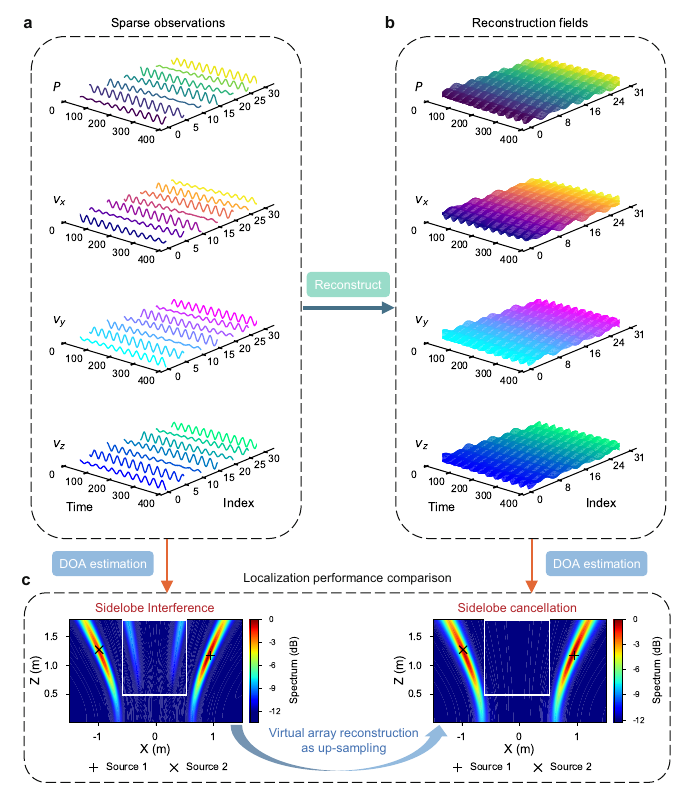}}
\caption{
        \textbf{Generative super-resolution for acoustic source localization.} In this experiment, two target sound sources emitted continuous sinusoidal signals at 2900 Hz and 3000 Hz. 
        \textbf{a}, Sparse observations from the 7-element physical AVS array, operating in a severely undersampled regime. 
        \textbf{b}, Reconstructed spatiotemporal dynamics across the 32-node virtual array, recovering the high-frequency wavefield continuous profile.
        \textbf{c}, DOA estimation via the full-vector MUSIC algorithm. The left panel shows the spatial spectrum derived from the original sparse physical array, where undersampling induces severe spatial aliasing and sidelobe interference (white box). The right panel shows the spatial spectrum based on the generative virtual array. By utilizing virtual array reconstruction as a spatial upsampling strategy, sidelobe elimination is successfully achieved. The plus ($+$) and cross ($\times$) symbols indicate the true locations of the sound sources.
    }
\label{fig:localization}
\end{figure}

In addition to recovering continuous physical fields, we demonstrate how the PI-ISS framework fundamentally redefines the hardware constraints of array signal processing by circumventing the spatial Nyquist limit. Conventional acoustic sensing systems are trapped in an inescapable physical trade-off: constraints on hardware costs and acquisition channels dictate a harsh compromise between array aperture, which governs spatial resolution, and spatial sampling density, that prevents aliasing. Consequently, expanding the physical aperture to resolve adjacent sources forces the inter-sensor spacing to exceed half the target signal wavelength. This severe spatial undersampling induces spatial aliasing and fundamentally confounds direction-of-arrival (DOA) estimation. Here, we demonstrate a paradigm shift from hardware-bound physical arrays to software-defined generative arrays. We show that the MRSM framework, coupled with PI-ISS sampling, acts as a physics-consistent spatial interpolator. It iteratively synthesizes dense virtual arrays from highly sparse physical measurements, effectively neutralizing the spatial undersampling bottleneck.

The free-field localization experiment was conducted in a semi-anechoic chamber to approximate reflection-free conditions (see Extended Data Figure \ref{fig:extended_data_2} for setup details). The measurement system comprised a uniform linear array of 7 in-house developed AVS units, each synchronously acquiring acoustic pressure and three-dimensional particle velocity. The physical inter-sensor spacing was fixed at 0.16 m. We introduced two independent sound sources emitting continuous sinusoidal signals at 2900 Hz and 3000 Hz. At these operating frequencies, the acoustic half-wavelength is approximately 0.058 m, significantly smaller than the physical sensor spacing, placing the measurement system in a severe state of spatial undersampling. 

We first constructed a full-vector free-field dataset via FDTD simulations to complete the MRSM pre-training phase. During inference, the sparse time-series observations, acquired by the 7-element physical array (Figure \ref{fig:localization}a) , were injected as conditional information into the reverse diffusion process. Guided by physical consistency, PI-ISS successfully reconstructed a continuous spatiotemporal physical field comprising 32 virtual nodes from the discrete observations (Figure \ref{fig:localization}b). This generative upsampling reduced the equivalent virtual inter-sensor spacing to 0.032 m, satisfying the spatial Nyquist sampling criterion across the entire field, and replenishing the critical waveform and phase information in the unobserved spatial regions.

To quantify the impact of this generative spatial upsampling, we executed the full-vector MUSIC spatial spectrum estimation algorithm on both the original physical array and the reconstructed virtual array. As depicted in Figure \ref{fig:localization}c (left), the spatial spectrum from the 7-element physical array exhibited spatial aliasing. It generated prominent grating lobes in the central interference zone with a peak spurious level of -8.25 dB, which severely confounded true target identification. In contrast, the 32-element virtual array generated by PI-ISS effectively increased the spatial sampling rate and neutralized these unphysical artifacts. Quantitative evaluation within this identical spatial zone demonstrates that the virtual array drives the local peak interference level down to -12.14 dB, a substantial absolute spatial interference suppression of 3.89 dB, successfully eliminating the sidelobe aliasing in the resulting localization spectrum (Figure \ref{fig:localization}c, right). Although the peak amplitude of the virtual array spectrum exhibited a slight attenuation due to minor reconstruction residuals, the structural fidelity of the localization was decisively preserved.

These experimental results conclusively demonstrate that integrating generative priors with physical constraints can successfully decouple array performance from spatial sampling density. By virtually reconstructing the unobserved wavefield continuity, the PI-ISS circumvents the spatial Nyquist limit without requiring additional physical sensors, which fundamentally revolutionizes the design paradigm of sensing systems. It proves that generative models, anchored by physical laws, can computationally substitute expensive hardware channels. Ultimately, this generative virtual array strategy holds out the prospect for deploying ultra-sparse, low-cost sensing architectures without sacrificing high-fidelity spatial awareness.

\subsection{Generalization to Kolmogorov flows}

\begin{figure}
\centering
\makebox[\textwidth][c]{\includegraphics[width=\textwidth]{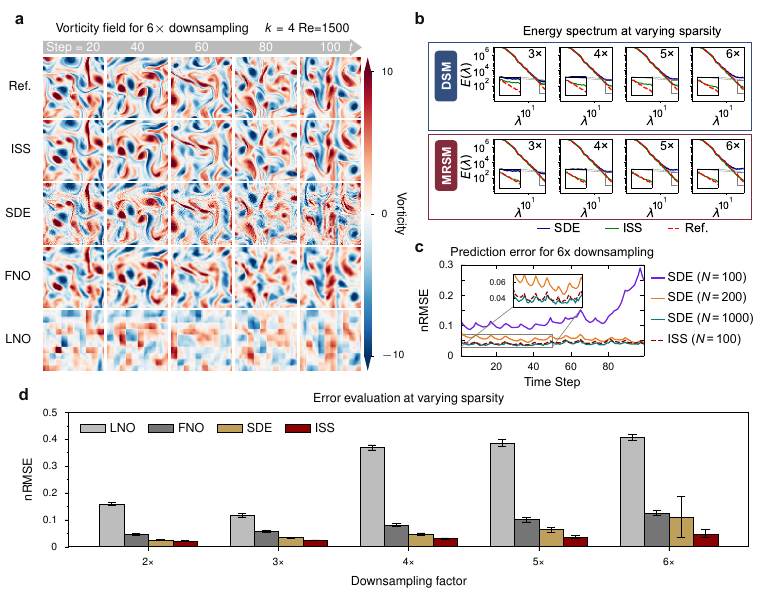}}
\caption{
        \textbf{Performance evaluation of Kolmogorov flow reconstruction at high Reynolds numbers.} 
        \textbf{a}, Reconstructed vorticity fields. The panel displays the results of the SDE, ISS, FNO, and LNO methods on a test sample with $k=4$ and Re = 1500, where the baseline operator networks were trained under $6\times$ spatial downsampling. 
        \textbf{b}, Impact of FPE regularization on the kinetic energy spectrum. The top and bottom rows present the energy spectra without and with FPE regularization, respectively. 
        \textbf{c}, Evolution of the nRMSE across 100 physical time steps. The spatiotemporal sequence generated by the proposed ISS (using $N=100$ reverse diffusion iterations) is evaluated against standard SDE implementations ($N=100, 200, 1000$). 
        \textbf{d}, Reconstruction performance across varying spatial downsampling factors. Data are presented as mean values $\pm$ standard deviation over 5 independent test samples with differing initial conditions.
    }
\label{fig:kolmogorov_flow}
\end{figure}

The preceding sections have established the exceptional efficacy of the PI-ISS framework in acoustic environments, where explicit governing equations could provide rigid gradient constraints to guide the generative trajectory. Real-world physical systems, however, often present a more severe challenge: governing equations may be partially unknown; crucial system parameters, such as the Reynolds number in turbulent flows, might be completely inaccessible during sparse observation. A truly universal generative framework must rely entirely on the intrinsic robustness of its learned prior to navigate ill-posed inversions and remain stable when explicit physical guidance is unavailable.

To validate this universality, we transition from explicitly constrained acoustics to the highly nonlinear and chaotic regime of Kolmogorov flows. This dynamic system is characterized by complex, multi-scale vortical structures that thoroughly test the capacity of purely data-driven generative priors. In the crucial part of the system, we deliberately deactivate the physics-informed gradient constraints, and deploy the foundational ISS driven solely by the FPE-regularized MRSM prior. We also challenge the framework with an out-of-distribution inference task at a higher, unseen Reynolds number (Re = 1500) and a distinct forcing wavenumber ($k=4$). This extreme configuration isolates and evaluates the pure representational capacity of the pre-trained prior. It demonstrates whether the MRSM framework merely relies on explicit equation penalization, or if it fundamentally internalizes the underlying physical manifold to achieve broad applicability across diverse dynamical systems.

Comparisons under a $6\times$ spatial downsampling constraint (Figure \ref{fig:kolmogorov_flow}a) reveal stark performance discrepancies. The end-to-end LNO fails to capture valid structures, and FNO exhibits severe distortions, whereas the standard SDE recovers macroscopic features but remains corrupted by high-frequency noise and artifacts. In contrast, the vorticity field reconstructed by ISS accurately reflects the complex vortex evolution, demonstrating robust feature extraction and high-fidelity completion from sparse data.

The physical consistency of these reconstructions is further validated in the frequency domain. Energy spectrum analysis (Figure \ref{fig:kolmogorov_flow}b) demonstrates that without FPE regularization, the purely data-driven model displays unphysical high-frequency energy fluctuations. This indicates that standard score networks generate non-conservative gradient fields in low-density regions. By enforcing the Score FPE during pre-training, the model mathematically guarantees the reverse martingale property, effectively eliminating these non-conservative components. Consequently, the energy spectrum of the FPE-regularized ISS reconstruction aligns exceptionally well with the reference ground truth, successfully restoring the true energy cascade across both the inertial subrange and the dissipation range.

Quantitative evaluation across the physical temporal evolution (Figure \ref{fig:kolmogorov_flow}c) highlights the computational superiority of the ISS sampling mechanism. After tracking the nRMSE across the 100 consecutive physical time steps of the generated spatiotemporal sample, we find that the proposed ISS, requiring only   reverse diffusion iterations, sustains a prediction error substantially lower than the standard SDE solver at equivalent or even 200-step configurations. Its accuracy over the entire temporal sequence closely rivals the computationally expensive 1000-step SDE baseline. This confirms that ISS reduces the inference cost by an order of magnitude without sacrificing the fidelity of long-term dynamic predictions. Finally, evaluating the models across progressively harsher downsampling factors (Figure \ref{fig:kolmogorov_flow}d) further underscores the resilience of this approach. While the errors of FNO and LNO escalate rapidly with increased sparsity, ISS sustains a persistently low error regime, exhibiting exceptional anti-perturbation capabilities when confronted with extreme observational constraints.

\subsection{Generalization to real-world meteorological systems}

\begin{figure}
\centering
\makebox[\textwidth][c]{\includegraphics[width=0.905\textwidth]{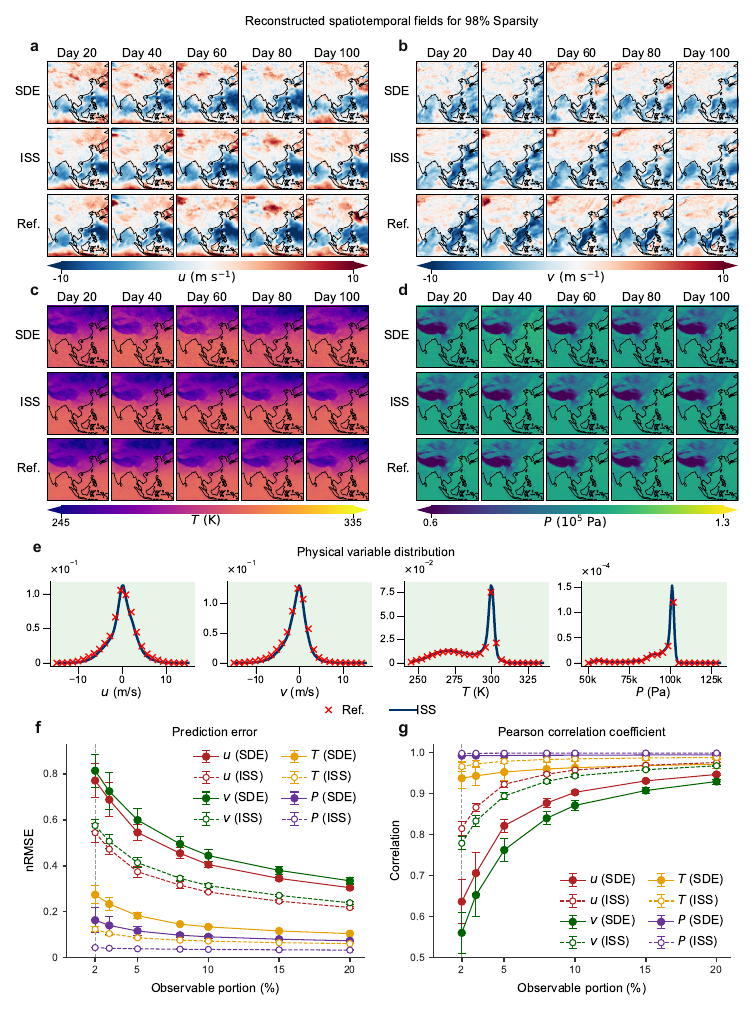}}
\caption{
        \textbf{Performance evaluation of ERA5 meteorological field reconstruction.} 
        \textbf{a–d}, Representative snapshots of the spatiotemporal reconstructions sampled at 20-day intervals over the first 100 days of 2023 under the 98\% data masking condition. The reverse diffusion process was constrained to 50 sampling steps across all results. \textbf{a}, 10-metre zonal wind velocity $u$. \textbf{b}, 10-metre meridional wind velocity $v$. \textbf{c}, 2-metre temperature $T$. \textbf{d}, Surface pressure $P$. 
        \textbf{e}, Statistical distributions displaying the PDFs of the four physical variables across all spatiotemporal points. 
        \textbf{f}, \textbf{g}, Quantitative evaluation as a function of the observation data proportion (ranging from 2\% to 20\%). Solid and dashed lines correspond to the ISS and standard SDE methods, respectively. Error bars represent the mean $\pm$ standard deviation across 5 independent random seeds. \textbf{f}, nRMSE. \textbf{g}, Pearson correlation coefficient.
    }
\label{fig:era5}
\end{figure}

We subject the framework to a stringent empirical evaluation: macro-scale, real-world Earth systems upon the universal representation capacity demonstrated in simulated chaotic flows. Meteorological dynamics, a formidable inverse challenge in the physical sciences, is characterized by profound multiphysics coupling, immense spatiotemporal dimensions, and inherently sparse observational networks. To evaluate whether the unconstrained ISS can navigate such complexity, we applied it to the East Asian domain of the ERA5 climate dataset to reconstruct highly coupled meteorological fields, namely, 10-metre zonal wind velocity ($u$), 10-metre meridional wind velocity ($v$), 2-metre temperature ($T$), and surface pressure ($P$).

Figure \ref{fig:era5} displays representative snapshots of the spatiotemporal reconstructions sampled at 20-day intervals over the first 100 days of 2023. During inference, only 2\% spatial grid points were selected randomly as conditioning inputs, representing an extreme 98\% missing data scenario. All generative trajectories were uniformly constrained to a rapid 50-step regime. Figures \ref{fig:era5}a–d contrast the reconstructions of the standard SDE solver against those of the proposed ISS across the four physical channels. While both solvers are driven by the identical pre-trained MRSM prior, the standard SDE solver suffers from severe local truncation errors under this accelerated, large-step sampling constraint, exhibiting noticeable local distortions and unphysical noise. Conversely, the ISS, leveraging its higher integration accuracy in the discrete update rule, mitigates these numerical bottlenecks and recovers coherent regional meteorological patterns. The ISS reconstructions align rigorously with the reference ground truth in critical macroscopic zones, such as the tropical dynamics over the lower latitudes and the thermal gradient structures at mid-latitudes, highlighting its capacity to preserve structural fidelity even with large sampling step sizes.

Figure \ref{fig:era5}e presents the probability density functions (PDFs) of the respective variables. The statistical distribution curves of the ISS-reconstructed fields (blue solid lines) are in precise agreement with the empirical ERA5 data (red crosses). This verifies that the ISS not only effectively suppresses point-wise reconstruction errors but also strictly reproduces the underlying physical and statistical distributions of the regional meteorological system.

Quantitative tracking of the reconstruction metrics (Figures \ref{fig:era5}f, g)  further delineates the performance hierarchy with the observable data proportion increaseing from 2\% to 20\%. The accuracy of ISS consistently surpasses that of the baseline SDE across the entire tested sparsity range. Crucially, under the most extreme sparsity (the 2\% observable ratio), the error of the SDE surges drastically, whereas the ISS sustains a persistently low nRMSE and high Pearson correlation. This underscores the exceptional robustness and inference efficiency of the ISS sampler in tackling severely ill-posed inverse problems in high-dimensional, nonlinear, and multiphysics-coupled physical systems.

\section{Discussion}

We reconstruct consistent dynamics by constraining generative priors with physical manifolds. MRSM embeds FPE regularization to guarantee stability, while PI-ISS projects stochastic trajectories onto deterministic manifolds using conservation residuals. This architecture effectively decouples prior learning from posterior physical correction, ensuring mechanical validity in accelerated sampling. Efficacy is validated across acoustic and chaotic systems. PI-ISS enables cross-modal inference and alleviates hardware constraints imposed by the spatial Nyquist criterion through virtual array synthesis. The stable prior, in ERA5 and Kolmogorov flows, preserves energy cascades and suppresses high-frequency artifacts. Consequently, the framework exhibits superior robustness compared to pure operator learning methods.

This superiority suggests a paradigm shift. Operator networks like FNO, mapping inputs to outputs directly, induce physical hallucinations under sparsity and therefore need complete retraining for new sensor geometries. Conversely, PI-ISS treats reconstruction as probabilistic sampling bounded by physics. Decoupling the data-driven prior from inference-time enforcement, without retraining, endows the model with zero-shot generalization capabilities, across highly variable and irregular sparse observation patterns. Furthermore, the decoupled structure of PI-ISS theoretically provides a foundation for future extensions into joint state-parameter inference. By leveraging generative priors to bridge observational gaps, this paradigm could potentially calibrate latent physical parameters within known PDE templates, a promising future investigation.

Despite these capabilities, fundamental challenges remain. First, iterative SDE-based sampling inherently trails the single forward-pass latency of operator networks. Although ISS significantly reduces the number of integration steps, meeting stringent real-time engineering requirements may still require future coupling with rapid-sampling distillation techniques. Second, the explicit gradient guidance in PI-ISS currently necessitates complete PDE formulations. Since residual convergence only confirms adherence to prescribed equations rather than underlying reality, zero residuals can inadvertently fit incomplete physics, introducing uncertainty in highly nonlinear regimes. This guidance also becomes intractable in environments with unknown parameters or fluctuating boundaries, thereby forcing reliance on the unconstrained ISS prior. While reconstructing global climate dynamics from merely 2\% of ERA5 observations within 50 steps demonstrates extreme resilience, reducing observation density below critical limits deprives the cross-gradient mechanism of spatial anchors, even causing mode collapse. Furthermore, extreme out-of-distribution noise from severe sensor failures can misdirect physical guidance. Therefore, a critical next step would be integrating learnable parameterizations to concurrently invert fields and unknown dynamics, alongside establishing theoretical sparsity bounds.

\section{Methods}

Our proposed generative framework is constructed hierarchically across three methodological layers. First, to establish a dynamically stable prior distribution, we introduce MRSM during the pre-training phase. Second, to overcome the computational bottleneck of standard reverse diffusion, we derive an ISS that serves as a highly efficient unconditional inference engine. Finally, for multiphysics-coupled systems, we extend this engine into a PI-ISS by embedding explicit physical governing laws into the sampling trajectory as gradient constraints.

\subsection{Martingale-regularized score matching (MRSM)}

Score-based generative models define a diffusion process indexed by a continuous time variable $t \in [0,T]$. Let $\mathbf{x}(t) \in \mathbb{R}^d$ represent the continuous state of high-dimensional data. Its stochastic dynamics evolving over time $t$ are governed by an Itô stochastic differential equation (SDE):
\begin{equation}\tag{1}
\mathrm{d}\mathbf{x} = \mathbf{f}(\mathbf{x}, t)\mathrm{d}t + g(t)\mathrm{d}\mathbf{w}
\end{equation}
where $\mathbf{w}$ is the standard Wiener process, $\mathbf{f}(\mathbf{x},t)$ is the drift coefficient, and $g(t)$ is the diffusion coefficient.

By initiating this process from a prior distribution $\mathbf{x}(T) \sim p_T$ and evolving backward in time, samples from the target distribution $\mathbf{x}(0) \sim p_0$ can be recovered. The corresponding reverse-time SDE \cite{andersonReversetimeDiffusionEquation1982} is expressed as:
\begin{equation}\tag{2}
\mathrm{d}\mathbf{x} = \left[ \mathbf{f}(\mathbf{x}, t) - g(t)^2 \nabla_{\mathbf{x}} \log p_t(\mathbf{x}) \right]\mathrm{d}t + g(t)\mathrm{d}\mathbf{\bar{w}}
\end{equation}
where $\mathbf{\bar{w}}$ is a standard Wiener process in the backward direction, and $\mathrm{d}t$ is an infinitesimal negative time step. Associated with Eq. (2) is a deterministic probability flow ordinary differential equation (PF-ODE) that shares the identical marginal probability density $p_t(\mathbf{x})$:
\begin{equation}\tag{3}
\mathrm{d}\mathbf{x} = \left[ \mathbf{f}(\mathbf{x}, t) - \frac{1}{2} g(t)^2 \nabla_{\mathbf{x}} \log p_t(\mathbf{x}) \right]\mathrm{d}t
\end{equation}

This work adopts the VE SDE framework, wherein the drift coefficient is $\mathbf{f}(\mathbf{x},t) = \mathbf{0}$ and the diffusion coefficient satisfies $g(t) = \sqrt{\mathrm{d}\sigma^2(t)/\mathrm{d}t}$.

In practice, a neural network $\mathbf{s}_\theta$ is trained to approximate the true score function $\nabla \log p_t$. To distinguish the spatiotemporal physical field from the generic state variable $\mathbf{x}$ utilized in standard generative models, we introduce the uppercase tensor $\mathbf{X}$ to represent the continuous physical field encompassing spatiotemporal correlations. The network $\mathbf{s}_\theta(\mathbf{X},t)$ is optimized via DSM \cite{songScoreBasedGenerativeModeling2021}:
\begin{equation}\tag{4}
\mathcal{L}_{DSM} = \mathbb{E}_{t, \mathbf{X}(0), \mathbf{X}(t)} \left[ \lambda(t) \left\| \mathbf{s}_\theta(\mathbf{X}(t), t) - \nabla_{\mathbf{X}(t)} \log p_{0t}(\mathbf{X}(t) | \mathbf{X}(0)) \right\|_2^2 \right]
\end{equation}
where $\lambda(t) > 0$ is a weighting function. Although DSM provides a standard training objective, the learned score function often deviates from the fundamental Score FPE, leading to non-conservative gradient fields and spatial instabilities \cite{laiFPDiffusionImprovingScorebased2023, laiEquivalenceConsistencyTypeModels2023a}.

To address this, we incorporate Score FPE regularization during the pre-training stage of the spatiotemporal physical fields. When the score field strictly satisfies the Score FPE, the absence of spurious non-conservative forces is mathematically guaranteed. Theoretically, this ensures the reverse diffusion trajectory adheres to the reverse martingale property, thereby establishing a dynamically stable generative prior for physical inverse problems. The MRSM training objective is defined as the sum of the standard DSM loss and the Score FPE regularization term:
\begin{equation}\tag{5}
\mathcal{J}(\theta) = \mathcal{L}_{DSM} + \gamma \mathcal{L}_{FPE}
\end{equation}

The regularization term $\mathcal{L}_{FPE}$ is defined as the squared norm of the residual operator $\mathcal{E}[\mathbf{s}_\theta]$, which quantifies the deviation of $\mathbf{s}_\theta$ from the Score FPE. For the VE SDE framework ($\mathbf{f} = \mathbf{0}$), this residual operator $\mathcal{E}[\mathbf{s}_\theta](\mathbf{X}, t)$ simplifies to:
\begin{equation}\tag{6}
\mathcal{E}[\mathbf{s}_\theta](\mathbf{X}, t) := \frac{\partial \mathbf{s}_\theta}{\partial t} - \nabla_{\mathbf{X}} \left( \frac{1}{2} g(t)^2 \mathrm{div}_{\mathbf{X}}(\mathbf{s}_\theta) + \frac{1}{2} g(t)^2 \|\mathbf{s}_\theta\|_2^2 \right)
\end{equation}

By jointly minimizing $\mathcal{J}(\theta)$, the pre-trained model is projected into a dynamical solution space which satisfies the martingale property. To ensure the computational tractability of the high-order derivatives in Eq. (6), the FDM is employed for the temporal derivative $\partial \mathbf{s}_\theta / \partial t$, and Hutchinson's trace estimator is utilized to stochastically approximate the divergence term $\mathrm{div}_{\mathbf{X}}(\mathbf{s}_\theta)$. This approach effectively balances numerical efficiency with physical autonomy.

\subsection{Implicit Score Sampler (ISS)}

Having established a dynamically stable generative prior via MRSM during the pre-training phase, we focus on the subsequent challenge in the inference stage. Conventionally, score-based generative models map noise to data, by solving the VE SDE in reverse; however, such standard numerical solvers are fundamentally constrained by efficiency bottlenecks. Classical numerical solutions (e.g., the Euler-Maruyama method) rely on the assumption of infinitesimal time steps. When the discrete time step is directly enlarged to accelerate sampling, the local truncation errors of the drift and diffusion terms will be amplified, causing the generated sample trajectories to deviate from the true data manifold and eventually resulting in reconstruction distortions. To circumvent this reliance on vanishingly small step sizes and to resolve the trade-off between sampling efficiency and generation quality, we derive an efficient sampling algorithm, termed the ISS, tailored specifically for the VE SDE framework.

The model corresponding to the VE SDE, under a discrete setting, is generally referred to as Score Matching with Langevin Dynamics (SMLD) \cite{songGenerativeModelingEstimating2020}. To establish the algorithmic foundation, we temporarily revert to the generic state variable $\mathbf{x} \in \mathbb{R}^d$. Considering a set of discrete noise scales $\{\sigma_k\}_{k=0}^N$ in SMLD (where the diffusion step index $k$ decreases from $N$ down to 0, satisfying $\sigma_N > \sigma_{N-1} > \dots > \sigma_0$), the joint probability distribution of this reverse process can be factorized as follows:
\begin{equation}\tag{7}
p_\theta(\mathbf{x}_{0:N}) = p(\mathbf{x}_N) \prod_{k=1}^N p_\theta(\mathbf{x}_{k-1} | \mathbf{x}_k)
\end{equation}
where $\mathbf{x}_{0:N}$ represents the complete evolution sequence from pure noise to data, $p(\mathbf{x}_N)$ is the prior noise distribution, and $p_\theta(\mathbf{x}_{k-1} | \mathbf{x}_k)$ is the reverse transition kernel. We define the reverse transition kernel at each step as a Gaussian distribution:
\begin{equation}\tag{8}
p_\theta(\mathbf{x}_{k-1} | \mathbf{x}_k) = \mathcal{N}\left(\mathbf{x}_{k-1}; \boldsymbol{\mu}_\theta(\mathbf{x}_k, \sigma_k), \eta^2(\sigma_k^2 - \sigma_{k-1}^2)\mathbf{I}\right)
\end{equation}
where $\boldsymbol{\mu}_\theta$ is the mean vector of the distribution, the variance is determined by a hyperparameter $\eta \in [0, 1]$ alongside the difference between adjacent noise variances, and $\mathbf{I}$ is the identity matrix. To determine $\boldsymbol{\mu}_\theta$, it is necessary to estimate the noise-free sample $\mathbf{\hat{x}}_0$ from the noisy sample $\mathbf{x}_k$ at the current diffusion step. Applying Tweedie's formula \cite{efronTweediesFormulaSelection2011} in conjunction with the score function $\mathbf{s}_\theta(\mathbf{x}_k, \sigma_k)$, the empirical Bayes estimate of the noise-free sample is obtained:
\begin{equation}\tag{9}
\mathbf{\hat{x}}_0 = \mathbf{x}_k + \sigma_k^2 \mathbf{s}_\theta(\mathbf{x}_k, \sigma_k)
\end{equation}

Upon acquiring the estimate $\mathbf{\hat{x}}_0$, the mean of the reverse transition is formulated as a linear combination of the current sample $\mathbf{x}_k$ and the predicted noise-free sample $\mathbf{\hat{x}}_0$:
\begin{equation}\tag{10}
\boldsymbol{\mu}_\theta(\mathbf{x}_k, \sigma_k) = \mathbf{x}_k + (\sigma_k - \sigma_{k-1})\sigma_k \mathbf{s}_\theta(\mathbf{x}_k, \sigma_k)
\end{equation}
This formulation corresponds to the deterministic component of the sampling process (i.e., the score-driven term), which leverages the score function to evolve the sample in the direction of increasing marginal probability density.

Added to the mean is a stochastic noise injection term scaled by the noise difference, so as to introduce controllable stochasticity during sampling and thereby enhance generation quality or correct prediction errors. Combining the score-driven term with the stochastic correction term yields an iterative rule for reverse evolution:
\begin{equation}\tag{11}
\mathbf{x}_{k-1} = \underbrace{\mathbf{x}_k + (\sigma_k - \sigma_{k-1})\sigma_k \mathbf{s}_\theta(\mathbf{x}_k, \sigma_k)}_{\text{Score-driven term}} + \underbrace{\eta\sqrt{\sigma_k^2 - \sigma_{k-1}^2}\mathbf{z}}_{\text{Stochastic correction term}}
\end{equation}
where $\mathbf{z} \sim \mathcal{N}(\mathbf{0}, \mathbf{I})$ is standard normal noise. When $\eta = 0$, the sampling process degenerates into a purely deterministic numerical solution of the PF-ODE; when $\eta > 0$, stochastic perturbation is introduced to augment diversity. The theoretical superiority of this sampler, namely, its linear exactness and first-order strong convergence bound under the variance-exploding regime, is mathematically established in Supplementary Note S1.

Regarding the high-dimensional sequence generation task inherent in spatiotemporal physical field reconstruction, the signals acquired by sensors typically span a prolonged physical duration $\tau$. Consequently, generating the sequence globally in a single pass poses severe challenges, including memory overflow and difficulties in capturing long-range dependencies. To resolve this, we reformulate the continuous dynamic physical field tensor $\mathbf{X}$ into a controllable spatiotemporal patching problem, and design a parallel generation strategy based on overlapping time windows. For the continuous spatiotemporal sub-sequences acquired by sparse sensors, we initialize $B$ continuous spatiotemporal noise tensor patches in the initial pure noise stage (diffusion step $k=N$):
\begin{equation}\tag{12}
\mathbf{X}_N = \left[\mathbf{X}_N^{(1)}, \mathbf{X}_N^{(2)}, \dots, \mathbf{X}_N^{(B)}\right]
\end{equation}
where the superscript $(b) \in \{1, 2, \dots, B\}$ denotes the parallel spatial patch index. Along the physical time dimension $\tau$, each patch tensor maintains an overlap of $m$ physical snapshots with its preceding adjacent patch. Subsequently, the ISS discrete iterative inference is executed independently for each patch. To ensure the physical temporal continuity of adjacent patches within the overlapping regions and to constrain the generated results to align with the sparse observations $\mathbf{y}^{(b)}$, conditional gradient guidance terms for observation consistency and temporal continuity are introduced during the discrete iteration (from step $k$ to $k-1$):
\begin{equation}\tag{13}
\begin{split}
\mathbf{X}_{k-1}^{(b)} &= \mathbf{X}_k^{(b)} + (\sigma_k - \sigma_{k-1})\sigma_k \mathbf{s}_\theta(\mathbf{X}_k^{(b)}, \sigma_k) + \eta\sqrt{\sigma_k^2 - \sigma_{k-1}^2}\mathbf{Z}^{(b)} \\
&\quad - \mathrm{clip}\left(\alpha \nabla_{\mathbf{X}_k^{(b)}} \| \mathbf{y}^{(b)} - \mathbf{H}\mathbf{\hat{X}}_0^{(b)} \|_2^2 + \beta \nabla_{\mathbf{X}_k^{(b)}} \mathcal{C}(\mathbf{\hat{X}}_0^{(b)}, \mathbf{\hat{X}}_0^{(b-1)})\right)
\end{split}
\end{equation}
where $\mathbf{Z}^{(b)} \sim \mathcal{N}(\mathbf{0}, \mathbf{I})$ is an independent standard Gaussian tensor corresponding to the $b$-th patch; $\mathbf{H}$ is the measurement mask matrix in the physical space; $\mathcal{C}(\cdot, \cdot)$ is the continuity loss function quantifying the discrepancy of physical snapshots within the overlapping regions; $\alpha$ and $\beta$ are hyperparameters modulating observation consistency and content similarity in the overlaps, respectively. Here, $\mathrm{clip}(\cdot)$ denotes a gradient clipping operation, utilized to constrain extreme conditional gradient values to prevent numerical instability. By iterating this process as the diffusion step $k$ decreases from $N$ down to 0, the parallel local sub-sequences are seamlessly fused, outputting the final denoised physical field.

\subsection{Physics-Informed Implicit Score Sampler (PI-ISS)}

The proposed ISS, though successfully resolves the inference efficiency problem, practically functions as a data-driven sampling engine. In complex spatiotemporal systems, relying solely on unconstrained statistical generation does not promise adherence of the resulting fields to the underlying physical laws. The spatiotemporal dynamics of continuous media are primarily governed by PDEs. For a generalized multimodal physical field $\mathbf{X}$ comprising $J$ components, the governing system is expressed as:
\begin{equation}\tag{14}
    \mathcal{P}_j(\mathbf{X}; \boldsymbol{\psi}) = \mathbf{0}, \quad j=1,\dots,J
\end{equation}
where $\mathcal{P}_j$ is a spatiotemporal differential operator and $\boldsymbol{\psi}$ represents the physical parameters of the medium. On a discrete computational grid, these continuous equations are transformed into residual forms. For the $j$-th physical conservation law, the discrete residual $\mathbf{R}_j$ at spatiotemporal coordinates $(\mathbf{r}, \tau)$ is defined as:
\begin{equation}\tag{15}
    \mathbf{R}_j(\mathbf{X}) := \widehat{\mathcal{P}}_j(\mathbf{X}; \boldsymbol{\psi})
\end{equation}
where $\widehat{\mathcal{P}}_j$ denotes the discrete physical operator formulated via the FDM or alternative numerical schemes.

Even though score-based generative models excel at fitting complex data distributions, purely data-driven generation cannot guarantee adherence to underlying physical laws. Theoretically, it is feasible to incorporate physical residuals, as a regularization term, into the MRSM pre-training loss, while it is an exorbitant computational burden to compute these residuals for every diffusion state during training.

To achieve efficient and physically consistent reconstruction, we adopted the Physics-Informed ISS (PI-ISS) to defer the imposition of physical constraints to the inference stage. At each discrete diffusion step $k$, we first estimate the noise-free physical field sample $\widehat{\mathbf{X}}_0^{(b)}$ from the current state $\mathbf{X}_k^{(b)}$ using Tweedie's formula: $\widehat{\mathbf{X}}_0^{(b)} = \mathbf{X}_k^{(b)} + \sigma_k^2 \mathbf{s}_\theta(\mathbf{X}_k^{(b)}, \sigma_k)$. Subsequently, a total physical potential energy function, $\mathcal{U}_{phys}$, is evaluated on this estimated noise-free field. Defined as the weighted sum of squared residuals across all coupled physical equations, this function quantifies the extent of physical violation:
\begin{equation}\tag{16}
    \mathcal{U}_{phys}(\widehat{\mathbf{X}}_0^{(b)}) = \sum_{j=1}^J \omega_j \left\| \mathbf{R}_j(\widehat{\mathbf{X}}_0^{(b)}) \right\|_2^2
\end{equation}
where $\omega_j$ is the weighting coefficient for the $j$-th equation. To synergistically generate multimodal fields, the negative gradient of this potential energy is integrated into the ISS update rule. Combined with the previously defined constraints for observation consistency and temporal continuity, the comprehensive iterative update rule for PI-ISS is formulated as:
\begin{equation}\tag{17}
    \mathbf{X}_{k-1}^{(b)} = \mathbf{X}_k^{(b)} + (\sigma_k - \sigma_{k-1})\sigma_k \mathbf{s}_\theta(\mathbf{X}_k^{(b)}, \sigma_k) + \eta \sqrt{\sigma_k^2 - \sigma_{k-1}^2} \mathbf{Z}^{(b)} - \mathbf{g}_{total}^{(b)}
\end{equation}

Here, the total guidance term $\mathbf{g}_{total}^{(b)}$ aggregates both the physical and observational constraints:
\begin{equation}\tag{18}
    \mathbf{g}_{total}^{(b)} = \mathrm{clip}\left( \alpha \nabla_{\mathbf{X}_k^{(b)}} \left\| \mathbf{y}^{(b)} - \mathbf{H}\widehat{\mathbf{X}}_0^{(b)} \right\|_2^2 + \beta \nabla_{\mathbf{X}_k^{(b)}} \mathcal{C}(\widehat{\mathbf{X}}_0^{(b)}, \widehat{\mathbf{X}}_0^{(b-1)}) + \lambda_{phys} \nabla_{\mathbf{X}_k^{(b)}} \mathcal{U}_{phys}(\widehat{\mathbf{X}}_0^{(b)}) \right)
\end{equation}
where $\lambda_{phys}$ is a hyperparameter governing the strength of the physical guidance. Under this update rule, the sampler not only gravitates towards regions of high probability density but is simultaneously projected towards the null space of the physical operator $\mathcal{P}$. We designate this ISS, enriched with physical constraints, as PI-ISS.

We prove the asymptotic physical consistency of the PI-ISS framework (detailed in Supplementary Note S2), which verifies that the physics-informed guidance mechanism drives the generated distribution toward the solution manifold defined by the physical conservation laws. Specifically, as the physical guidance strength approaches infinity ($\lambda_{phys} \to \infty$), the generated sample sequence converges weakly to the exact intersection of the data prior support and the physical manifold.

This theoretical validity demonstrates that PI-ISS fundamentally constructs a physics-constrained Gibbs distribution. Through a cross-gradient coupling mechanism, PI-ISS constrains the multimodal variables to evolve jointly within a physically admissible solution space. For the acoustic scenarios emphasized in this paper, this ensures that acoustic pressure and particle velocity, coupled via the momentum equation, will not undergo unphysical independent evolution.

In practical applications of the PI-ISS algorithm, a finite physical guidance strength $\lambda_{phys}$ numerically corresponds to a soft projection. Although Theorem 2 establishes asymptotic consistency under limiting conditions, the actual reconstruction error remains constrained by the discretization step size of the sampling algorithm and the specific choice of $\lambda_{phys}$.

\subsection{Implementation of PI-ISS in acoustic scenarios}

This section elucidates PI-ISS framework’s specific implementation in reconstructing spatiotemporal acoustic field. We designed physical constraint mechanisms, tailored to diverse sensor configurations, for both unimodal and multimodal scenarios.

\textbf{Unimodal constraints.} In acoustic pressure field reconstruction solely relying on microphone arrays, the target tensor $\mathbf{X}$ exclusively comprises the acoustic pressure component $p$ on a discrete spatial grid. The generated spatiotemporal samples are constrained to satisfy the second-order wave equation in linear acoustics. For an isotropic and homogeneous medium, this physical constraint operator $\mathcal{P}_1$ is defined as:
\begin{equation}\tag{19}
    \mathcal{P}_1(p; c) = \nabla^2 p - \frac{1}{c^2} \frac{\partial^2 p}{\partial \tau^2} = 0
\end{equation}
where $\nabla^2$ denotes the Laplace operator. In the discrete computational domain, the second-order central FDM is employed to construct the discrete residual $\mathbf{R}_1$. For a spatiotemporal grid point $(\mathbf{r}, \tau)$, the discrete physical loss term $\mathcal{U}_{scalar}$ is expressed as:
\begin{equation}\tag{20}
    \mathcal{U}_{scalar}(\widehat{\mathbf{X}}_0^{(b)}) = \left\| \delta_{\mathbf{r}}^2 \hat{p}_0^{(b)} - \frac{1}{c^2} \delta_{\tau}^2 \hat{p}_0^{(b)} \right\|_2^2
\end{equation}
where $\delta_{\mathbf{r}}^2$ and $\delta_{\tau}^2$ represent the spatial and temporal discrete Laplacian difference operators, respectively. This loss term constrains the spatiotemporal evolution of the generated acoustic pressure field to satisfy wave consistency.

\textbf{Multimodal synergistic constraints.} In reconstruction tasks utilizing AVS, the target tensor $\mathbf{X}$ constitutes a multimodal composite field comprising the acoustic pressure $p$ and the particle velocity vector $\mathbf{v} = [v_x, v_y, v_z]^T$. To achieve the physically entangled generation of these modalities, we introduce multi-level physical consistency constraints, encompassing coupled equations describing inter-modal correlations and self-constraint equations governing individual modalities.

The Euler equations constrain the acoustic pressure gradient and the time derivative of the particle velocity:
\begin{equation}\tag{21}
    \mathbf{R}_{mom}(\mathbf{X}) := \rho_0 \frac{\partial \mathbf{v}}{\partial \tau} + \nabla p = \mathbf{0}
\end{equation}
Simultaneously, the continuity equation constrains the temporal evolution of acoustic pressure and the divergence of particle velocity:
\begin{equation}\tag{22}
    R_{cont}(\mathbf{X}) := \frac{\partial p}{\partial \tau} + \rho_0 c^2 \nabla \cdot \mathbf{v} = 0
\end{equation}
Both acoustic pressure and particle velocity individually satisfy the wave equation in a homogeneous medium. Unimodal self-constraint terms arefurther introduced, to ensure each physical component maintains correct spatiotemporal wave characteristics:
\begin{equation}\tag{23}
\begin{aligned}
    R_{wave, p}(\mathbf{X}) &:= \nabla^2 p - \frac{1}{c^2} \frac{\partial^2 p}{\partial \tau^2} = 0 \\
    \mathbf{R}_{wave, \mathbf{v}}(\mathbf{X}) &:= \nabla^2 \mathbf{v} - \frac{1}{c^2} \frac{\partial^2 \mathbf{v}}{\partial \tau^2} = \mathbf{0}
\end{aligned}
\end{equation}
For multimodal scenarios, the total physical energy function $\mathcal{U}_{vector}$ is formulated as the weighted sum of squares of all aforementioned coupled residuals and self-constraint residuals:
\begin{equation}\tag{24}
    \mathcal{U}_{vector}(\widehat{\mathbf{X}}_0^{(b)}) = \omega_{mom} \| \mathbf{R}_{mom} \|_2^2 + \omega_{cont} \| R_{cont} \|_2^2 + \omega_p \| R_{wave, p} \|_2^2 + \omega_{\mathbf{v}} \| \mathbf{R}_{wave, \mathbf{v}} \|_2^2
\end{equation}

Note that Eq. (24) serves as the concrete realization of the generalized PI-ISS constraint defined in Eq. (16), specifically tailored for the entangled generation of acoustic pressure and particle velocity fields.

The weighting coefficients $\{\omega_j\}$ dictates the efficacy of the physical guidance. Given the dimensional disparity between acoustic pressure (unit: Pa) and particle velocity (unit: m/s), their corresponding residual values are imbalanced by orders of magnitude. Thus, a normalization process is conducted:

\textbf{Magnitude alignment based on characteristic impedance.} In an air medium, acoustic pressure and particle velocity are correlated via the characteristic impedance $Z_0 = \rho_0 c$. The residual magnitude of the Euler equation is predominantly governed by the pressure gradient $\nabla p$, whereas the continuity equation contains a $\rho_0 c^2$ term, whose value is typically substantially higher than the former. By establishing the weight relationship $\omega_{mom} \approx Z_0^2 \cdot \omega_{cont}$, we ensure that the cross-modal coupling possesses equal optimization weight during the generation process.

\textbf{Consistency normalization of residuals.} The wave equation self-constraint term includes a spatial second-order derivative $\nabla^2$, whose magnitude is inversely proportional to the square of the grid step size $\Delta r$. To prevent the self-constraint terms from either overwhelming or being overshadowed by the coupling terms, a dynamic magnitude alignment strategy is adopted, defined as $\omega_j \propto \mathbb{E}\left[\| \mathbf{R}_j \|_2^2\right]^{-1}$.

Detailed elaborations on the multimodal cross-gradient coupling mechanism for acoustics are provided in Supplementary Note S3.

\section{Code Availability}

The code for MRSM is open source, released under the MIT License. The code repository is accessible online, at: \\ \href{https://github.com/BrianZhu1999/MRSM}{\color{blue}{https://github.com/BrianZhu1999/MRSM}}.

\section{Data Availability}

The training and testing datasets, and the pre-trained model weights for the MRSM framework, are openly available in the Zenodo repository at: \\ \href{https://doi.org/10.5281/zenodo.19398315}{\color{blue}{https://doi.org/10.5281/zenodo.19398315}}.

\section{Acknowledgements}

This work was supported in part by the Projects of Jiangsu Province Science and Technology Plan Special Fund in 2023, Basic Research Program Natural Science Foundation, under Grant BK20232048; and in part by the National Key Research and Development Program of China under Grant 2022YFB2602003 and Grant 2023YFA1406904.

\section{Author Contributions}

M.B., Z.C. and M.L. supervised the project. M.B., Z.Z. and K.H. conceived the idea. Z.Z., M.B. and M.L. ensured the physical rigor of the framework. Z.Z., K.H. and M.B. formalized the mathematical rigor of the theories. Z.Z. and Z.C. performed the numerical simulations. Z.C., J.Z. and J.L. developed the AVS. Z.Z., Y.S. and Z.C. designed and carried out the experiments. G.W. provided guidance on computational mechanics modeling. H.X. provided guidance on physics-informed modeling. Q.Z. provided guidance on the analysis of Kolmogorov flows. X.L. and Y.C. provided scientific oversight and critical feedback. All authors discussed the results and contributed to manuscript preparation.

\bibliographystyle{naturemag}
\bibliography{references}

\renewcommand{\figurename}{Extended Data Figure}
\setcounter{figure}{0}

\begin{figure}[p]
\centering
\makebox[\textwidth][c]{\includegraphics[width=0.9\textwidth]{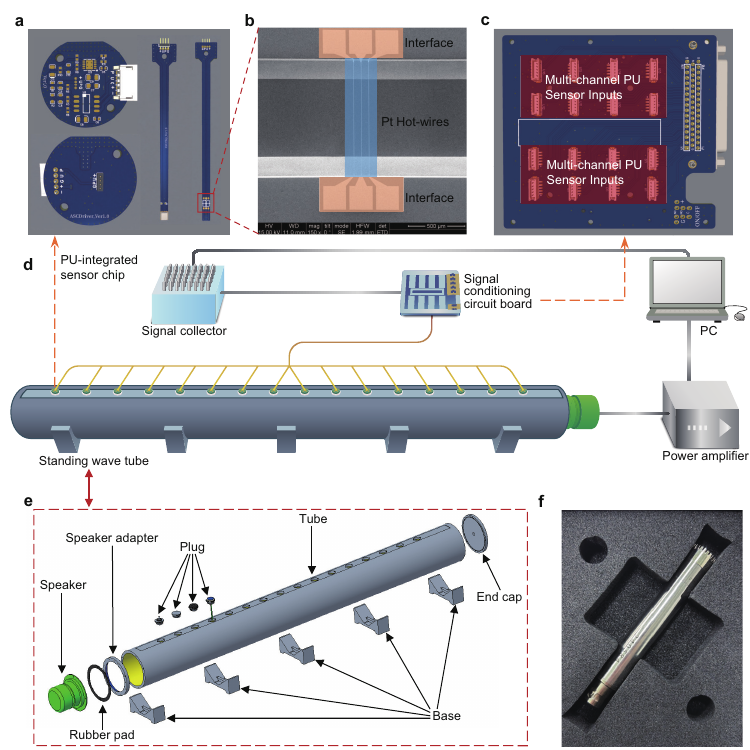}}
\caption{
        \textbf{Schematic of the data acquisition and processing system for the standing wave tube acoustic field.} 
        \textbf{a}, Design schematic of the PU joint acquisition sensor chip. The chip integrates measurement functionalities for both acoustic pressure and particle velocity, utilized for data acquisition in multimodal acoustic field reconstruction experiments. 
        \textbf{b}, Microscopic magnified view of the particle velocity sensing module. Shows microstructural details such as the platinum (Pt) hot-wires and the electrical interface. 
        \textbf{c}, In-house developed multi-channel signal conditioning circuit board. Used for the pre-processing and amplification of analog signals output by the sensors. 
        \textbf{d}, Experimental hardware connection setup. The PC emits a signal, which is amplified by a power amplifier to drive the loudspeaker, generating an acoustic field within the standing wave tube. The sensor array distributed along the tube wall acquires the acoustic field signals. After being processed by the signal conditioning circuit board, the signals are transmitted to the signal collector for aggregation and then fed back to the PC for analysis. 
        \textbf{e}, Structural schematic of the standing wave tube. The main body of the standing wave tube is fabricated from acrylic glass, with a total length of 1500 mm, a diameter of 150 mm, and a wall thickness of 10 mm. A loudspeaker is mounted at one end of the tube, while the other end is sealed with an end cap. Sixteen sensor mounting holes are reserved axially along the tube wall; the first measurement point is located 150 mm from the sound source, and the spacing between all subsequent adjacent points is maintained at 80 mm. All connecting parts are equipped with sealing measures to prevent sound leakage. 
        \textbf{f}, The 1/2-inch acoustic pressure sensor. Used for the scalar acoustic pressure field reconstruction experiment.
    }
\label{fig:extended_data_1}
\end{figure}

\begin{figure}[p]
\centering
\makebox[\textwidth][c]{\includegraphics[width=\textwidth]{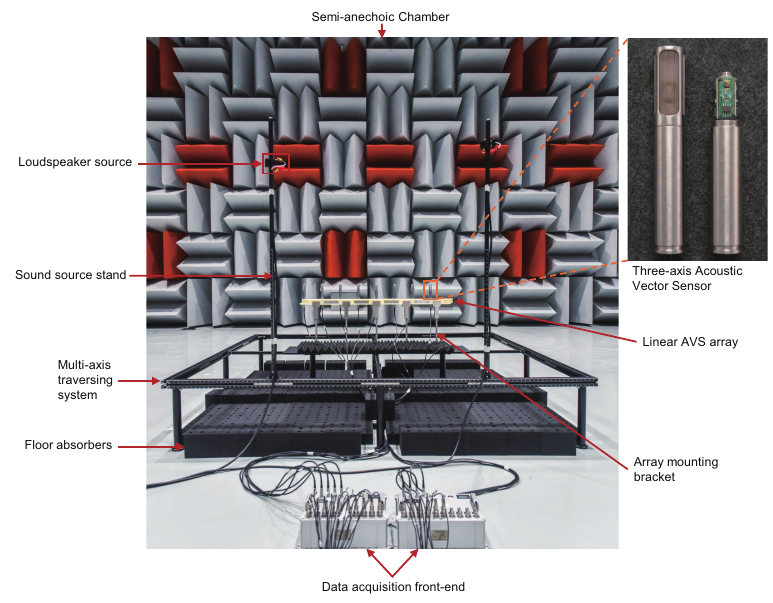}}
\caption{
        \textbf{Hardware and scene setup for the full-vector acoustic field localization experiment.} The experiment was conducted in a semi-anechoic chamber. The internal dimensions of the semi-anechoic chamber are 12.6 × 10.9 × 6.1 m$^3$, with a cut-off frequency of 80 Hz, a free-field radius of 4.5 m, and a background noise level below 1.3 dB. Acoustic absorbing foam was laid on the floor to attenuate ground reflections. A multi-axis translation stage and array fixtures ensured the spatial precision of the AVS linear array and the loudspeaker sources on both sides. The photograph shows a representative 8-element AVS arrangement on the same experimental platform; the localization experiment reported in this work used a 7-element AVS linear array with an inter-sensor spacing of 0.16 m. The top-right corner of the figure provides a close-up of the external appearance and internal structure of the in-house developed three-axis AVS used in the array. The multi-channel data acquired by the sensors were synchronously recorded through the data acquisition system front-end at the bottom.
    }
\label{fig:extended_data_2}
\end{figure}

\end{document}